\documentclass{article} 
\usepackage{iclr2016_conference,times}
\usepackage{hyperref}
\usepackage{url}
\usepackage{graphicx}
\usepackage{subcaption}
\usepackage{import}
\usepackage{algorithm}
\usepackage{algorithmic}
\usepackage{amsfonts}
\usepackage{mathtools}
\usepackage{array}
\usepackage{longtable}
\usepackage{booktabs}	
\graphicspath{{fig/}}

\title{Sequential Action-Induced Invariant Representation for Reinforcement Learning}

\author{
\hspace{-1mm}Dayang Liang,
Qihang Chen,
Yunlong Liu\footnotemark[1]\thanks{Corresponding Author}\\
Xiamen University \\
Xiamen, China\\
\texttt{ylliu @ xmu.edu.cn}
}

\iclrfinalcopy 

\begin{document}

\maketitle

\begin{abstract}
How to accurately learn task-relevant state representations from high-dimensional observations with visual distractions is a realistic and challenging problem in visual reinforcement learning. Recently, unsupervised representation learning methods based on bisimulation metrics, contrast, prediction, and reconstruction have shown the ability for task-relevant information extraction. However, due to the lack of appropriate mechanisms for the extraction of task information in the prediction, contrast, and reconstruction-related approaches and the limitations of bisimulation-related methods in domains with sparse rewards, it is still difficult for these methods to be effectively extended to environments with distractions. To alleviate these problems, in the paper, the action sequences, which contain task-intensive signals, are incorporated into representation learning. Specifically, we propose a \underline{S}equential \underline{A}ction--induced invariant \underline{R}epresentation (SAR) method, in which the encoder is optimized by an auxiliary learner to only preserve the components that follow the control signals of sequential actions, so the agent can be induced to learn the robust representation against distractions. We conduct extensive experiments on the DeepMind Control suite tasks with distractions while achieving the best performance over strong baselines. We also demonstrate the effectiveness of our method at disregarding task-irrelevant information by deploying SAR to real-world CARLA-based autonomous driving with natural distractions. Finally, we provide the analysis results of generalization drawn from the generalization decay and t-SNE visualization. Code and demo videos are available at \url{https://github.com/DMU-XMU/SAR.git}.
\end{abstract}

\section{Introduction}

Visual Deep Reinforcement Learning (DRL) with high-dimensional images as input can easily deal with decision-making problems in various complex scenarios, which has achieved great success in robot control \cite{Li2021ReinforcementLF}\cite{Kalashnikov2018QTOptSD}, autonomous driving \cite{Zhao2022CADREAC}\cite{Wu2022TrajectoryguidedCP}, video games \cite{jaderberg2019human}, health care \cite{liang2022treatment}, among others. However, observation signals in real-world applications are usually high-rank and unstructured, and the direct use of such high-dimensional input for visual DRL often results in information redundancy and sample inefficiency.

In the literature, many techniques have been adopted to learn the representation efficiently for visual DRL, and the main approaches are to learn a state representation jointly in an end-to-end manner with the help of convolutional encoder \cite{mnih2015human}, attention mechanisms \cite{liang2021gated} and Transformer \cite{vaswani2017attention}. Although these methods perform well in some low-rank high-dimensional backgrounds, it is still difficult to handle images from real-world applications that contain many task-irrelevant signals, where the sample cost for the learning of the policy in such scenarios will be very high. Fortunately, the number of basic states that directly guide the decision may be much lower than that of the image \cite{scholkopf2022causality}, which allows the representation learner to extract background-independent task states from complex observations, so that agents can be generalized in environments with similar tasks but different backgrounds.

Recently, data augmentation and encoder reconstruction have been widely employed to address the insufficient generalization performance of visual DRL. For example, self-supervised visual transformer \cite{dosovitskiy2020image} is used to reconstruct masked hidden states (MLR) \cite{Yu2022MaskbasedLR}, autoencoders (MAE) \cite{liumasked}\cite{he2022masked}, world models (MWM) \cite{seo2022masked} to promote the encoder to learn task-relevant features and predict current and future states.  Random data augmentation has also been widely used in sample augmentation (RAD) \cite{laskin2020reinforcement}, noise contrastive estimation (CURL) \cite{laskin2020curl}, and robust value estimation (DrQ) \cite{yarats2021image}. However, naively introducing data augmentation to arbitrarily transform observations and reconstruct encoders in a task-agnostic manner may cause problems of high variance and overfitting, further deteriorating the generalization performance \cite{yuan2022don} and leading to additional computational costs. 

By inducing representation with behavioral similarity metrics (BSM)~\cite{zang2022simsr}, recent research has achieved some success in task-relevant representation learning, where an element of the Markov decision process (MDP) tuple that is able to determine the task, such as the reward, action, or transition model, is adopted to establish the task equivalence relationship with the hidden state to enhance the learning of the lossless representation of task-relevant information. Related works mainly include: CRESP \cite{Yang2022LearningTR} based on sequence rewards, contrastive learning method \cite{liu2020return} based on MC-returns, DBC \cite{zhang2020learning} based on rewards and transitions, and PSM \cite{agarwal2020contrastive} based on one-step action and transition.

However, existing methods still have some limitations. For the reward-based metric methods, the limitations mainly include: i)  It is inaccurate to directly group observations with the equal one-step reward into a class of states. To distinguish different encoded states, the \textbf{trajectories with different rewards} are required, while only the one-step reward is adopted in current related approaches \cite{kemertas2021towards}. ii) In environments with sparse rewards, insufficient reward signals may directly lead to the failure of reward-based behavioral similarity metrics, thereby worsening representation learning \cite{agarwal2020contrastive} \cite{agarwal2021behavior}. For the action-based approaches, e.g., the method with PSM \cite{agarwal2020contrastive}, although more task-relevant information, i.e., the action, is considered in the representation learning, it is still not enough to learn an accurate representation with a one-step metric (the issue is similar to i)).

\begin{figure}[t]
\centering
\includegraphics[height=6cm]{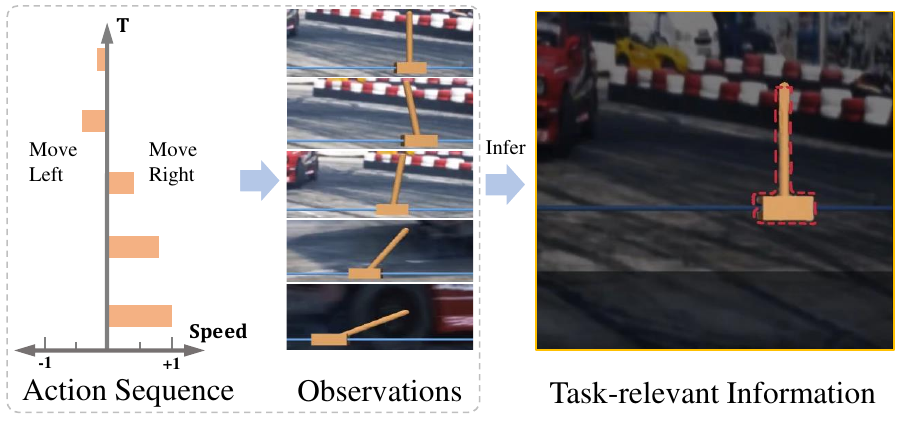}
\caption{We illustrate the idea of extracting task-relevant information using action sequences on a simple control task. Taking Cartpole-swingup as an example, its task is to control the slider to move left and right through the learned policy to finally achieve the balance of the connecting rod. 
We can see that with the navigation of the agent in the scenario, the movement rule of the regions~(\textbf{Middle}: slider and connecting rod) navigated by the agent are consistent with the sequential action signals (\textbf{Left}: action sequence), then if we can lock these background-irrelevant task regions/information~(\textbf{Right}: red highlight part) by those action signals, a more accurate task-relevant representation can be obtained for decision making. } 
\label{fig1}
\end{figure}

To address the aforementioned issues, the \textbf{action sequence distribution} is modeled and adopted for task-relevant representation learning with the purpose of decoupling task-related states and distractions accurately. This method stems from a crucial insight (further illustrated with Fig \ref{fig1}): task-related regions in a static observation are hard to recognize without prior but they can be moved by interactive actions, and further, sequential multi-step actions can cause them to move obviously, i.e., generate motion trajectories. Intuitively, these regions, i.e., task-relevant information can be efficiently recognized and located by multi-step actions. With this insight, we can model the idea of multi-step actions to accurately represent task-relevant information while avoiding the problem of inaccurate single-step metrics and unstable metrics of sparse reward.


In this paper, we propose a \underline{S}equential \underline{A}ction-induced invariant-\underline{R}epresentation (SAR) method, which makes the output of state encoder consistent with the control rules of real sequential actions (i.e., label) by minimizing the distance between the predicted and the real action sequence, thereby locking into task-relevant representation from the stacked observations. To reach the above idea, we utilize the characteristic function to model a probability distribution for real action sequences while leveraging the latest policy to guarantee the accuracy of real actions, thereby achieving an effective metric between real and predicted action sequences. Empirical evaluations on the DeepMind Control (DMControl) suite with unseen distractions demonstrate our approach outperforms recent representation-related baselines. To further verify the effectiveness of our approach, SAR is also applied to the field of autonomous driving with real natural distractions, experimental results show that significant performance than baselines can still be achieved. In addition, the results of the generalization decay ratio and t-SNE visualization analysis also strongly confirm that the method using sequential actions can extract task-relevant information more accurately.

\section{Background}

We model the underlying system of visual reinforcement learning as a Markov decision process (MDP), which can be described by a tuple $\cal M=(S, A,P, R,\gamma)$, where $\cal S$ represents the state space, $\cal A$ represents the continuous action space, ${\cal P}(s'|s, a){:\cal S} \times {\cal A} \times{\cal S} \to \mathbb{R}$ represents the transition probability from state $s$ to next state $s'$ after executing action $a\in \cal A$, ${\cal R}(s,a):{\cal S} \times {\cal A} \to \mathbb{R}$ represents the reward signal $r$ obtained after executing action $a$  under state $s$, $\gamma \in[0,1]$ represents the discount factor. Generally, the accurate state $s$ is difficult to obtain in environments with visual distractions. It is necessary to extract the hidden state from the observation $o \in \cal O$ with the help of state encoder $\phi$, where $\cal O$ is the observation space with task-irrelevant information.

The aim of visual reinforcement learning is to jointly learn an efficient encoder $\phi$ for state representation and a policy $\pi$ for decision. They allow the agent to extract task-relevant low-dimensional hidden state $\phi(o):o\to z$ from a given high-dimensional $o \sim \cal O$, and choose an action $a \sim \pi(z)$ according to the $z \in \mathbb{R}^n$, thus obtaining the reward signal $r={\cal R}(s, a)$ from the environment. By iterating the above process, the agent finally learns an optimal policy that maximizes the expected cumulative discounted rewards, formulated as $max_{\pi}\mathbb{E}_{s_t\sim {\cal P},a_t\sim \pi}[{\sum}^{\infty}_{t=0}{\gamma}^t {\cal R}(s_t,a_t)]$.
\subsection{Soft Actor Critic}
SAC is an off-policy reinforcement learning algorithm for continuous control, which consists of a Critic network for learning the value function $Q_{\varphi}(o; a)$ with parameters ${\varphi}$ and an Actor network for learning the policy function $\pi_{\psi}(a|o)$ with parameters ${\psi}$. Unlike value-based methods, SAC optimizes a stochastic policy to maximize the expected trajectory reward. The main highlight of SAC is that the maximum entropy item ${\cal H}=log(\pi_{\psi}(a|o))$ is added to decentralize sampling actions, which enhances the exploration and robustness of the algorithm.

Specifically, The Critic network first samples transition $e_t=(o_t,a_t,o_{t+1},r_t,d)$ from the replay buffer $\cal D$ to minimize the Bellman error, where $d$ represents done signal. The training of parameters $\varphi$ can be represented by the following loss:
\begin{equation}
\label{eq1}
{\cal L}^{V}(\varphi)=\mathbb{E}_{e\sim \cal D}\left[{\left(Q_{\varphi}(o_t,a_t)-(r_t+\gamma (1-d){\cal T})\right)}^2\right].
\end{equation}

The target $\cal T$ computes the expectation of the next actions sampling from current policy,  defined as:
\begin{equation}
\label{eq2}
{\cal T}=\mathbb{E}_{a'\sim \pi}\left[(\hat Q_{\hat{\varphi}}(o_{t+1},a')-\alpha log(\pi_{\psi}(a'|o_{t+1}))\right],
\end{equation}
where network $\hat{\varphi}$ comes from the Exponential Moving Average (EMA) of the Critic parameter $\varphi$, and the parameter $\alpha$ is a positive entropy coefficient that determines the priority of entropy maximization over value function optimization.

Finally, we sample actions $a\sim\pi_\psi$ from the policy with parameter $\psi$, and train the Actor by maximizing the expected reward of the sampled actions:

\begin{equation}
\label{eq3}
{\cal L}^{\pi}(\psi)=\mathbb{E}_{a\sim \pi}\left[(Q^{\pi}(o,a)-\alpha log(\pi_{\psi}(a|o))\right].
\end{equation}
	
\subsection{Behavior Similarity Metric}

Behavior Similarity Metrics~(BSM) related methods generally employ MDP tuples, such as reward, action, policy, etc., to measure the task equivalence between observations, where an auxiliary loss is usually constructed to promote the extraction of task-relevant state representations. Policy Similarity Metric (PSM) is a BSM-based method that utilizes policy distance. Since our work is closely related to PSM in terms of optimization, we briefly introduce the basic definition of PSM and the differences between them.

For the same point, both our method and the PSM method are based on the BSM principle, i.e., if the encoder distance is equal to the distance between certain MDP tuples (such as the PSM distance in Theorem 3.1), it is considered that all the information extracted by the encoder is task-related. For differences, PSM method optimizes the one-step distance $\lVert z_i-z_j \rVert_1$ of the encoded states, making it close to the one-step distance between the MDP tuples that can determine the task, i.e., the sum distance of the policy distance $DIST(\pi^* (x), \pi^* (y))$ with the next transition distance ${\cal W}_1$. However, in our method, besides the basic idea of sequence actions, the most important difference is that we minimize the difficult multi-step distance, while there is no need to learn the transition model ${\cal P}$ avoiding the impact of model accuracy.

\newtheorem{thm}{\bf Theorem}[section]
\begin{thm}\label{thm1}
(Policy Similarity Metric \cite{agarwal2020contrastive}). \it{Let $\mathfrak{M}$ be the set of all pseudometrics on space $S$. For given DIST and ${\pi}^*$, A pseudometric transformation function $F_{PSM}^{{\pi}^*}: \mathfrak{ met\to met}$ is defined as,}
\begin{equation}
\label{eq4}
{\cal F}_{PSM}^{{\pi}^*} = DIST({\pi}^*(s_i),{\pi}^*(s_j))+\gamma {\cal W}_1(d^*)({\cal P}^{\pi^*}_{s_i},{\cal P}^{\pi^*}_{s_j}),
\end{equation}
\it{where DIST denotes a probability pseudometric between policies, $\pi^*$ is the optimal policy, transition ${\cal P}_s^{\pi}={\sum}_a\pi(a\vert s){\cal P}(\cdot \vert s,a)$ and ${\cal W}_1$ is the 1-Wassersrein distance given the pseudometric $d$. Then, ${\cal F}_{PSM}^{{\pi}^*}$ has a unique fixed point $d^*$ which is a PSM metric.}
\end{thm} 

\subsection{Distraction Background}
As SAR  uses sequential actions for the distinction of the task-relevant representation and the distraction, it is required that the task foreground and the distraction background should have a low coupling property, i.e., their movements do not exhibit a consistent relationship, which actually holds in many real-world applications. Since in reality,  the distraction backgrounds of most scenes are random and not coupled with task foregrounds, also it is really difficult for multi-degree-of-freedom systems to couple with distractions, thus our method is applicable to the state representation of most scenes with distractions. Nevertheless, we still give the following assumptions.
\vspace{5pt}
\newtheorem{assumption}{\bf Assumption}[section]
\begin{assumption}\label{assp1}
\it{Given a reinforcement learning system, where the transition model of states is denoted as ${\cal P}_S$ and the transition model of distraction background is denoted as ${\cal P}_X$, then we assume that $D_{KL} ({\cal P}_S \parallel {\cal P}_X )\gg 0$, i.e., the similarity of the two distributions are nearly to zero.}
\end{assumption}

\section{Algorithm}

In this section, we propose Sequential Action-induced invariant Representation~(SAR), a visual DRL method that leverages optimal action sequences to capture task-relevant information from sequential observations. The key detail of SAR is to optimize the potential distance between the encoding state and the real sequential actions through an auxiliary loss to make the prediction vector of the encoding state consistent with the control signal of the sequential action. In this way, SAR is able to promote the encoding hidden state to only preserve the components that follow the control signals of real sequential actions,  thereby ensuring that the representation extracted from the observation is task-relevant. 

\begin{figure*}[h]
\centering
\includegraphics[height=6.3cm]{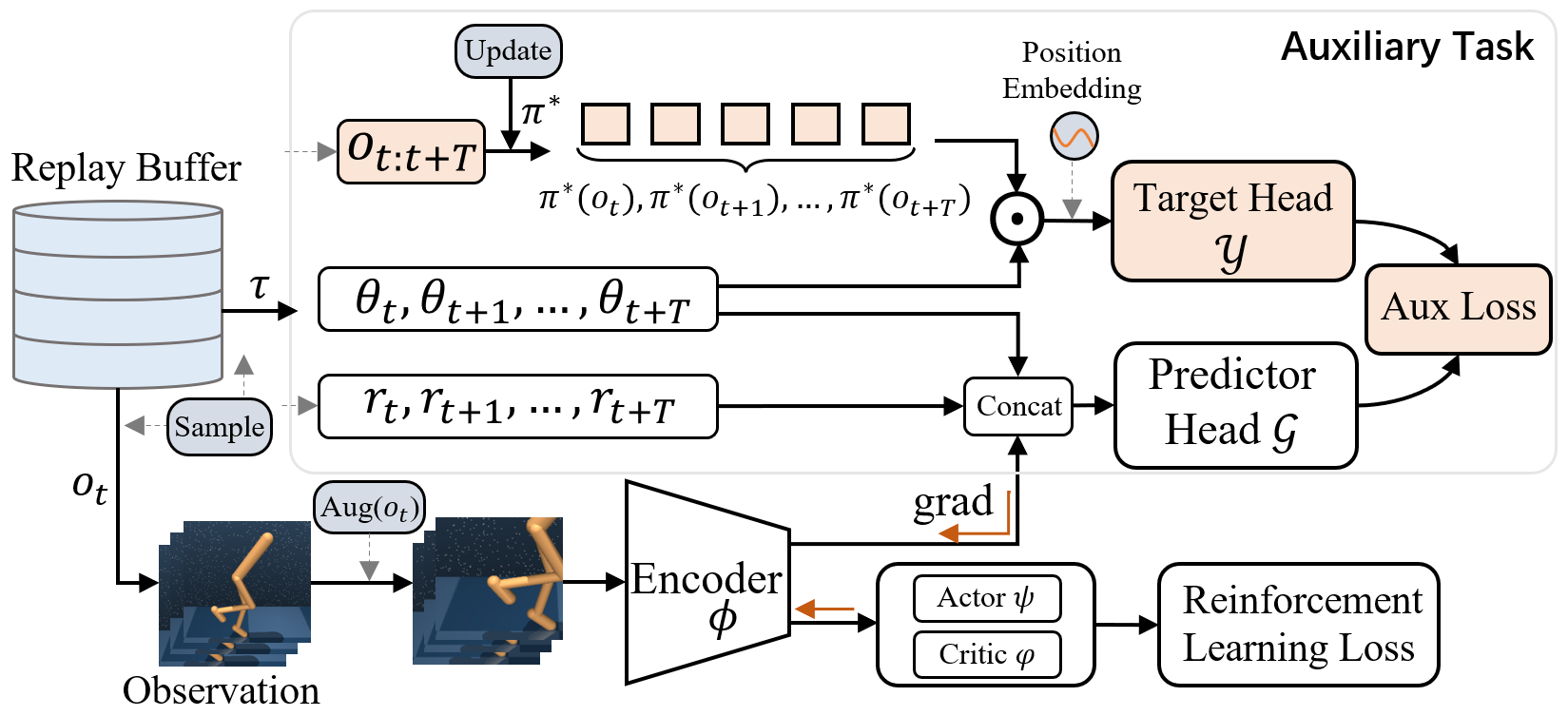}
\caption{Overview architecture of SAR. SAR consists of two parts, an auxiliary task for learning representations and a reinforcement learning task for learning policies. The box region is the core module, i.e., the auxiliary task, which aims to minimize the error between the predicted output signal $\cal G$ and the optimal action sequence distribution $\cal Y$. Thus, the gradient of auxiliary loss can be back-propagated to $\phi$, assisting $\phi$ to learn state representations associated with real action sequences from unstructured scenes. Finally, the efficient representation learned by the encoder will be worked for the decision task of reinforcement learning.} 
\label{fig2}
\end{figure*}

The overall framework of SAR is shown in Fig. \ref{fig2}, which is based on the Soft Actor-Critic reinforcement learning architecture. In detail, SAR consists of an auxiliary module~(Aux Loss) for self-supervised representation learning and an RL module~(RL loss) for policy learning, where we mainly focus on the auxiliary module. In the architecture, the two modules share an encoder $\phi$ for state representation but affect it differently. Specifically, in the phase of backward gradient update, the RL module will train both the encoder network $\phi$ and the Actor-Critic network, while the auxiliary module will only train $\phi$. During forward inference, the pipeline of the auxiliary module is ignored, and only the RL module is used to obtain current action. In addition, to improve the generalization of the encoder network, we also randomly augment the observations before encoding according to existing methods.

\subsection{Construction of Auxiliary Task}
The construction method of auxiliary tasks tremendously determines the expected performance. Therefore, during the construction of auxiliary tasks, a key step is how to construct auxiliary objectives that can efficiently optimize the state encoder. To that end, we construct an auxiliary task with optimal action sequences and make them guide the extraction of task-relevant information. In detail, in order to take full advantage of the properties possessed by the action sequence, we establish a probability model for sequence actions and finally employ the characteristic function of the probability distribution as the target of the auxiliary loss. Before this implementation, we first use the newest policy to calculate the actions of sampling observations and try to find better action sequences.
\\

\textbf{Actions Sequence}~Sequential action-induced representation aims to lock task-relevant information in observations that have a specific mapping relationship with action sequences, so action sequences will directly guide the extraction of task-relevant information. In theory (Theorem 3.1), SAR should also utilize the optimal action sequences sampled from optimal policy $\pi^*$, which helps to learn representations accurately \cite{agarwal2020contrastive}. As for reason, we empirically believe that due to the optimal sequential actions sampled from an optimal probability distribution function (PDF), the information controlled by optimal actions is deeply decoupled from the random noise information. In contrast, information controlled by random actions cannot be accurately extracted due to its coupling with random noises.

In practice, the optimal actions cannot be obtained during training, but the actions sampled from the suboptimal policy can also be worked in representation learning with an accuracy limitation \cite{agarwal2020contrastive}. Therefore, to further improve the optimality of action sequences, we make the following improvement: At each training step, for the action $a_t$ sampled from the replay buffer, we do not directly use the $a_t$ that was calculated with old policy $\pi^{old}$ like Agarwal et al. Instead, we update the $a_t$ by the latest current policy. To be specific, we first sample sequential observations $\{o_t,o_{t+1},...,o_{t+T}\}$ of length $T$ from replay buffer, and then use the current policy $\pi^{new}$ to calculate the action sequence with the sampled sequential observations:
\begin{equation}
\label{eq5}
a^*_t,a^*_{t+1},...,a^*_{t+T} \approx \pi^{new}(o_t),\pi^{new}(o_{t+1}),...,\pi^{new}(o_{t+T}).
\end{equation}

Compared with the old policy adopted by Agarwal et al., current policy $\pi^{new}$ is thought to be closer to the optimal distribution of $\pi^*$ (see Assumption 4.1), so adopting the $\pi^{new}$ will guarantee a tighter lower bound of the action optimality. For easier reading, suboptimal policy $\pi^{new}$ and suboptimal action $a^{new}$ ($a^{new} \sim \pi^{new}(\cdot)$) will be marked with $\pi^*$ and $a^*$ respectively below.

\begin{assumption}\label{assp2}
\it{Given a reinforcement learning process for steady training. Before the policy $\pi$ converges, we empirically assume that the distribution of the current $\pi^{new}$ will be closer to the distribution of the optimal  $\pi^*$ as the number of gradient update steps increases, i.e., $D_{KL} (\pi^{new} \parallel \pi^* )<D_{KL}(\pi^{old} \parallel \pi^*)$.}
\end{assumption} 

\textbf{Characteristic Functions of Actions Sequence Distributions}~We now model the probability distribution of action sequences to promote the construction of the auxiliary loss target. It is worth noting that since the PDF of the action sequence is difficult to handle, we use the characteristic function of the action sequence distribution to replace the PDF according to the work of Yang et al. \cite{Yang2022LearningTR}. The following Lemma 4.1 shows that the characteristic function is a probability method that has an equivalent relationship with the PDF.
\vspace{5pt}
\newtheorem{lemma}{\bf Lemma}[section]
\begin{lemma}\label{lemma1}
[21] \it{Two random vectors $X$ and $Y$ have the same characteristic function if and only if they have the same probability distribution function.}
\end{lemma}
\vspace{5pt}

Next, the characteristic function of action sequence distribution is used as the target $\cal{Y}$ of the auxiliary loss in the modeling process and calculation formula.  

Consider a set of sequential actions $a^s=(a^*_t,a^*_{t+1},...,a^*_{t+T})\in\Psi$ under $\pi^*$, where $\Psi$ is the combined space of the T-step actions defined on original action space $\cal A$, and the probability distribution function of $a^s$ is expressed as $f_{\Psi}(a^s)$. Then, the characteristic function defined under $\Psi$ is expressed as follows:

\begin{align}
\label{eq6}
{\cal Y}_{\Psi}(\theta,a^s) &=\mathbb{E}_{a^s\sim f_{\Psi}}\left[e^{i<\theta,a^s>}\right] \\
&=\int e^{i<\theta,a^s>}f_{\Psi}(a^s)da^s.
\end{align}

Where imaginary unit $i=\sqrt{-1}$, inner product $<\theta,a^s>={\sum}^{k=t+T}_{k=t}\theta_k a^*_k$, and $\cal Y$ is the complex function of the real variable $\theta \sim \Theta$. $\Theta$ is the probability density function on $\mathbb{R}^T$, where Gaussian distribution $\Theta={\cal N}(\mu,\sigma^2)$ is actually adopted.

Finally, the characteristic function of action sequence distribution can be expressed by the real and imaginary units as follows:
\begin{equation}
\label{eq7}
{\cal Y}^{cos}=cos\left({\cal Y}_\Psi(\theta,a^s)\right),~{\cal Y}^{sin}=sin\left({\cal Y}_\Psi(\theta,a^s)\right).
\end{equation}
\subsection{Optimization Implementation}
In this subsection, we mainly introduce the implementation and optimization details of the auxiliary task and give the overall training algorithm of the SAR framework.

First, we estimate the characteristic function of the action sequence distribution using the neural network predictor $Pred$. To improve the generalization of representation learning, the input of the $Pred$ consists of a concatenation of encoding state $z$, real variable $\theta$, and reward sequence $r^s=\{r_t,r_{t+1},..,r_{t+T}\}$. Therefore, the output of $Pred$ prediction can be denoted as: 
\begin{equation}
\label{eq8}
{{\cal G}_{\phi} (o,\theta,r^s)}=Pred(\left[\phi(o),\theta,r^s\right]).
\end{equation}

Second, we compute the $L_2$ distance (mean squared error) between the predicted characteristic function and the true characteristic function of the action sequence, then optimizing the encoder $\phi$ through preliminary loss function:
\begin{align}
\label{eq9}
&{\cal L}^{MSE}({\cal G_\phi,Y}) \\
&=\mathbb{E}_{<o,r^s,a^s>\sim D,\theta\sim\Theta} \left\lVert {{\cal G}_\phi(o,\theta,r^s)}-{\cal Y}_{\Psi}(\theta,a^s)\right\rVert_2^2.
\end{align}

\begin{algorithm}[t]
    \caption{Sequential Action-Induced Invariant Representation}
    \label{algorithm1}
    \textbf{Input}: replay buffer $D$ with size $N$, learning rate, batch size, etc. \\
    \textbf{Output}: optimal $\pi$
    \begin{algorithmic}[1] 
    		\STATE Initialize Critic network $\varphi$, Actor network $\psi$, and encoder network $\phi$.
		\FOR {episode $m \gets 0$ to $M$}
			\STATE Encode state $z_t=\phi(o_t)$.
			\STATE Excute action $a_t=\pi(\cdot|z_t)$.
			\STATE Collect data $D\gets D\cup \{o_t,a_t,t_{t+1}\}$. 
		\ENDFOR
		\FOR{gradient step $i \gets 0$ to $I$}
			\STATE Sample batch $B_i\sim D$.
			\STATE Get sequence $\{a_{k:k+T},r_{k:k+T}\}$ via $B_i$.
			\STATE Update action sequence $a_{k:k+T}=\pi(o_{k:k+T})$.
			\STATE Train the Actor-Critic ${\cal L}^{V}+{\cal L}^{\pi}$. (Eq. \ref{eq1} and \ref{eq3}).
			\STATE Train the auxiliary task ${\cal L}^{MSE}+{\cal L}^{CS}$. (Eq. \ref{eq11} and \ref{eq12}).
		\ENDFOR
        \STATE \textbf{return} optimal $\pi$
    \end{algorithmic}
\end{algorithm}

In fact, since the true characteristic function cannot be directly obtained, following Yang et al. \cite{Yang2022LearningTR}, we optimize an upper bound of ${\cal L}^{MSE} (\cal G,Y)$:
\begin{align}
\label{eq10}
&{\cal L}^{MSE}({\cal G_\phi,Y}) \\
&=\mathbb{E}_{<o,r^s,a^s>\sim D,\theta\sim\Theta} \left\lVert {{\cal G}_\phi(o,\theta,r^s)}-e^{i<\theta,a^s>}\right\rVert_2^2 \\
&\ge\mathbb{E}_{<o,r^s>\sim D,\theta\sim\Theta} \left\lVert {{\cal G}_\phi(o,\theta,r^s)}-\mathbb{E}_{a^s\sim f_{\Psi}}\left[e^{i<\theta,a^s>}\right]\right\rVert_2^2
\end{align}

In Eq.\ref{eq10}, since the true characteristic function is represented by real and imaginary units, the predicted vector is also decoupled into two corresponding parts $[{\cal G}_\phi^{cos},{\cal G}_\phi^{sin}]=\cal G_\phi$. Finally, the loss function with decoupled real and imaginary parts is expressed as:
\begin{align}
\label{eq11}
&{\cal L}^{MSE}({\cal G_{\phi},Y}) \\
&=\mathbb{E}_{<o,r^s,a^s>\sim D,\theta\sim\Theta} \left[\lVert{\cal G}_\phi^{cos}-{\cal Y}^{cos}\rVert_2^2 +\lVert{\cal G}_\phi^{sin}-{\cal Y}^{sin}\rVert_2^2\right].
\end{align}

Additionally, we construct a cosine similarity (CS) loss function on the cosine component between the true characteristic and predicted vector. Unlike $L_2$ distance, which optimizes the magnitude of the prediction vector, the cosine similarity loss can guarantee the phase distance between the predicted and the target vector, thus making the prediction vector closest to the characteristic vector of the true action sequence.
\begin{align}
\label{eq12}
{\cal L}^{CS}&=Cosine({\cal G}_\phi^{cos},{\cal Y}^{cos}) \\
&=1-\frac{1}{k}\sum^{K-1}_{i=0}\frac{{\cal G}^{cos}_{\phi, i}}{\lVert{\cal G}^{cos}_{\phi, i}\lVert}\frac{{\cal Y}^{cos}_{i}}{\lVert{\cal Y}^{cos}_{i}\lVert}.
\end{align}

In the end, the above auxiliary representation learning module with ${\cal L}^{MSE}$ and  ${\cal L}^{CS}$ will be extended to the Soft Actor-Critic (SAC) reinforcement learning framework. And the AC framework includes the Actor loss  ${\cal L}^{\pi}$ for the training policy and the Critic loss  ${\cal L}^{V}$ for the training value network. Therefore, we train SAR through minimizing the total loss ${\cal L}={\cal L}^{MSE}+{\cal L}^{CS}+{\cal L}^{\pi}+{\cal L}^{V}$, and corresponding pseudocode can be seen in \textbf{Algorithm \ref{algorithm1}}.

\section{Results}

To effectively verify the generalization performance of our SAR and recent baselines under distracting environments, the main experiments set different distraction sources for the training and evaluation phases, i.e., the evaluation phase uses video backgrounds not seen during training. Then we report the comparison results of SAR and baselines in the challenging DeepMind Control (DMControl) suite tasks, and comprehensively analyze the generalization ability of SAR with the generalization decay ratio and t-SNE visualization. Finally, we also deploy the algorithms to the field of autonomous driving with real natural distractions to verify the superior performance. 

\subsection{Evaluation Setting}
\textbf{DMControl with Background Distraction} The DMControl suite~\cite{Tassa2018DeepMindCS} is a challenging visual environment based on the Mujoco physics simulator, which is widely used for the performance verification of visual reinforcement learning. Following to the distracting DMControl setting~\cite{stone2021distracting}, the background distraction of the experiment is sampled from natural videos in the DAVIS 2017 dataset. Specifically, algorithms use 2 videos alternately as background during the training phase and evaluation on unseen 30 videos. We adopt the above challenging settings to evaluate the anti-distraction representation ability of algorithms.

\textbf{Baselines} We compare SAR against recent strong visual RL baselines under the DMC suite: (i) CRESP~\cite{Yang2022LearningTR}, an algorithm that proposes to induce task-relevant state representations through reward sequences, is one of the current algorithms with outstanding generalization performances; (ii) DrQ~\cite{yarats2021image}, a recent state-of-the-art algorithm that utilizes data augmentation techniques to learn robust representations from pixels; (iii) DBC~\cite{zhang2020learning} which is a common RL baseline for learning compact encodings based on the task equivalence principle under bisimulation metric; (iiii) CURL~\cite{laskin2020curl}, a contrastive unsupervised representation learning method for RL, which has achieved state-of-the-art data-efficiency on clean pixel-based environments.

\subsection{Main Results}
\begin{figure*}[tb]
\centering
\includegraphics[height=8.3cm]{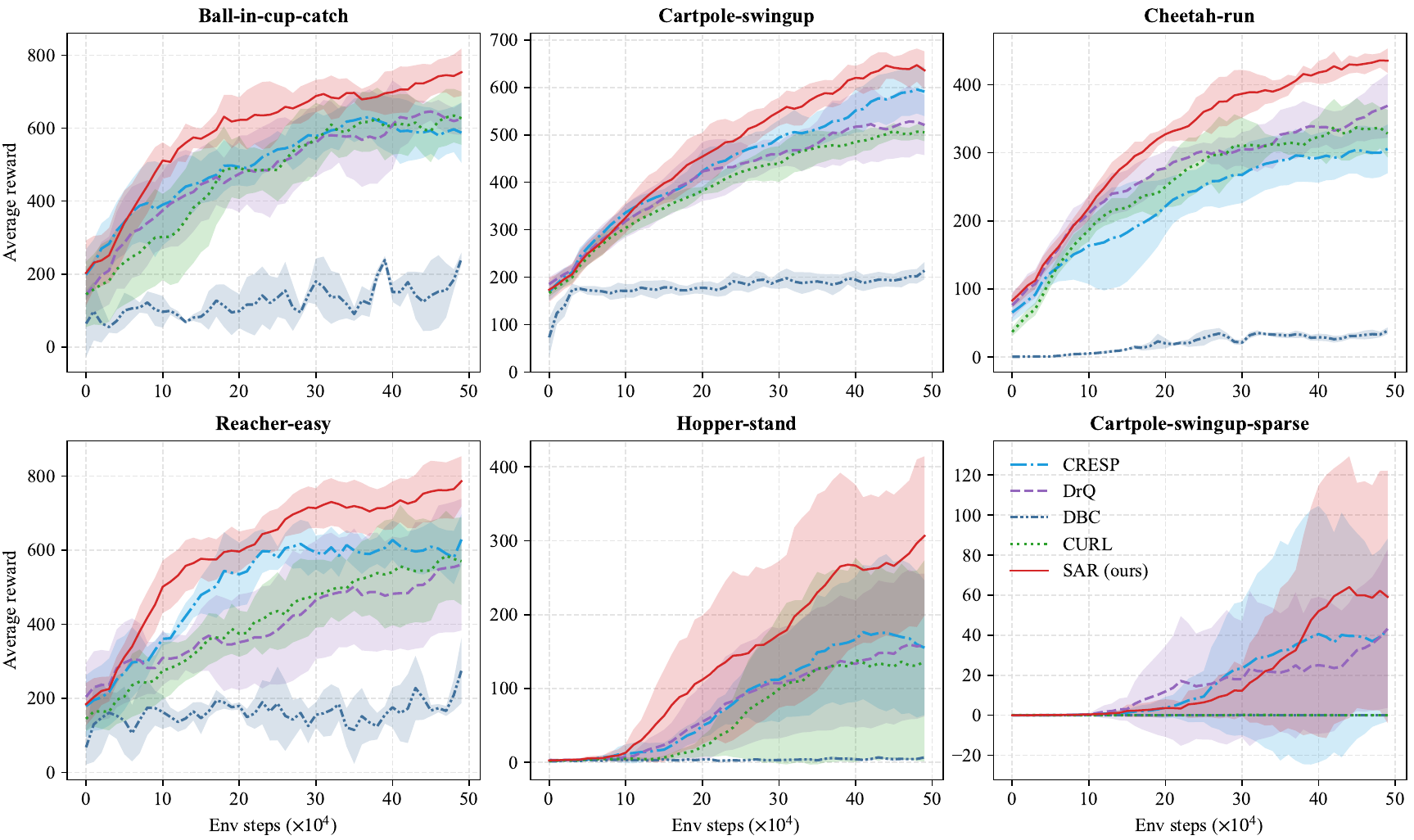}
\caption{The evaluation results for 500K environment steps on unseen natural video background settings in DMC tasks. Each method is trained using 5 different seeds; for every seed the mean episode returns are computed every 10000 environment steps, averaging over 10 episodes. All the data of baselines in this experiment come from the recurring results of the corresponding standard algorithms.}
\label{fig:3}
\end{figure*}

The evaluation results of the experiment on the DMControl tasks with the unseen distracting background are shown in Fig~\ref{fig:3}. In general, we evaluate the performance of SAR by comparing with four common baselines: DrQ, DBC, CURL, as well as the latest CRESP, in 6 tasks of different difficulty of Cartpole, Cheetah, Hopper, and Ball\_in\_up in four scenarios. In these experiments, all baselines adopt the settings consistent with the original work. For SAR, the learning rate is 5e-4, the initial temperature is 0.1, the update frequency of Critic and Actor is 2, and we use Adam optimizer \cite{kingma2014adam} to optimize the auxiliary task, Critic and Actor network with a batch size of 256. See Table \ref{tab:2} for the full of hyperparameters.

\begin{table*}[h]\scriptsize
    \centering
    \caption{Comparison results of the best episode scores on 6 DMC tasks.(mean $\&$ standard deviation for 5 seeds)}
    \label{tab:1}
    \begin{tabular}{lcccccc}
        \toprule
        Methods      & Bic-catch & Cartpole-swingup & Cheetah-Run & Reacher-easy & Hopper-stand & Cartpole-swingup-sp \\
        \midrule
        CURL         & 659$\pm$84 & 509$\pm$22 & 348$\pm$47 & 589$\pm$183 & 140$\pm$135 & 0$\pm$0 \\
        DBC          & 193$\pm$10 & 214$\pm$17 & 43$\pm$7 & 210$\pm$15 & 6$\pm$1 & 0$\pm$0 \\
        DrQ          & 659$\pm$22 & 533$\pm$67 & 379$\pm$57 & 573$\pm$161 & 159$\pm$96 & 43$\pm$40 \\
        CRESP        & 638$\pm$65 & 593$\pm$49 & 311$\pm$42 & 672$\pm$71 & 185$\pm$108 & 43$\pm$46 \\
        SAR (ours)   & \textbf{768$\pm$66} & \textbf{651$\pm$34} & \textbf{446$\pm$23} & \textbf{795$\pm$71} & \textbf{313$\pm$117} & \textbf{67$\pm$70} \\
        \bottomrule
    \end{tabular}
\end{table*}

In order to highlight that our method has a stronger representation ability for eliminating background distractions, we choose 4 difficult tasks (Ball-in-cup-catch, Cartpole-swingup, Cheetah-run, and Reacher-easy) from the DMC500K benchmark, and additional 2 high-difficulty tasks with sparse reward property (Hopper-stand, Cartpole-swingup-sparse). As shown in Table \ref{tab:1} based on 5 repeated experiments, the SAR method with sequential action-induced task information representation achieves the best performances on these tasks compared to the baselines.

\subsection{Probing the Generalization of Representations}

\begin{figure}[th]
\centering
\includegraphics[height=8cm]{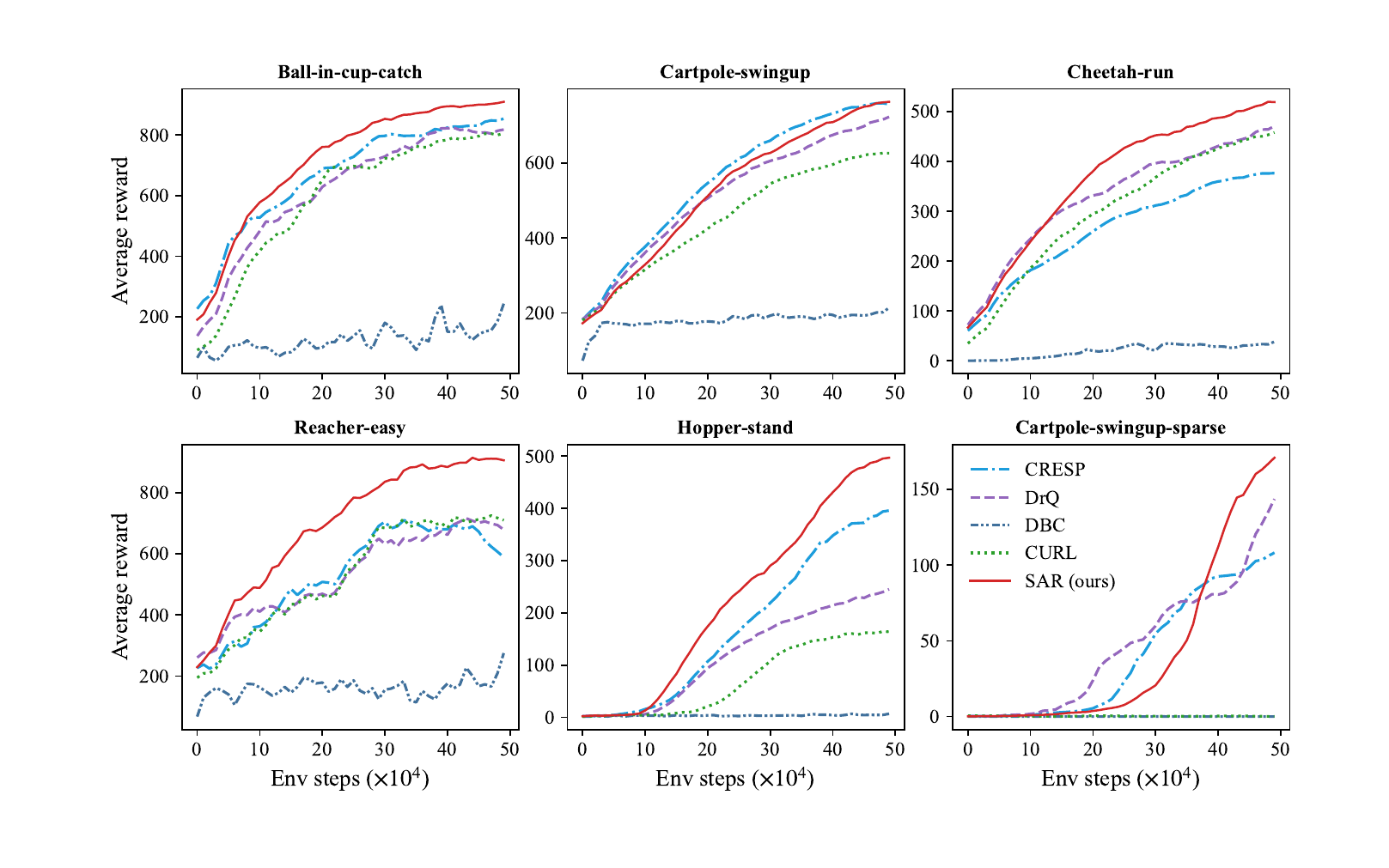}
\caption{The performance comparison curves of the models during the training stage with the seen distracting backgrounds. Results for each curve are averaged from 5 seeds.}
\label{fig:4}
\end{figure}
\begin{figure}[th]
\centering
\includegraphics[height=7cm]{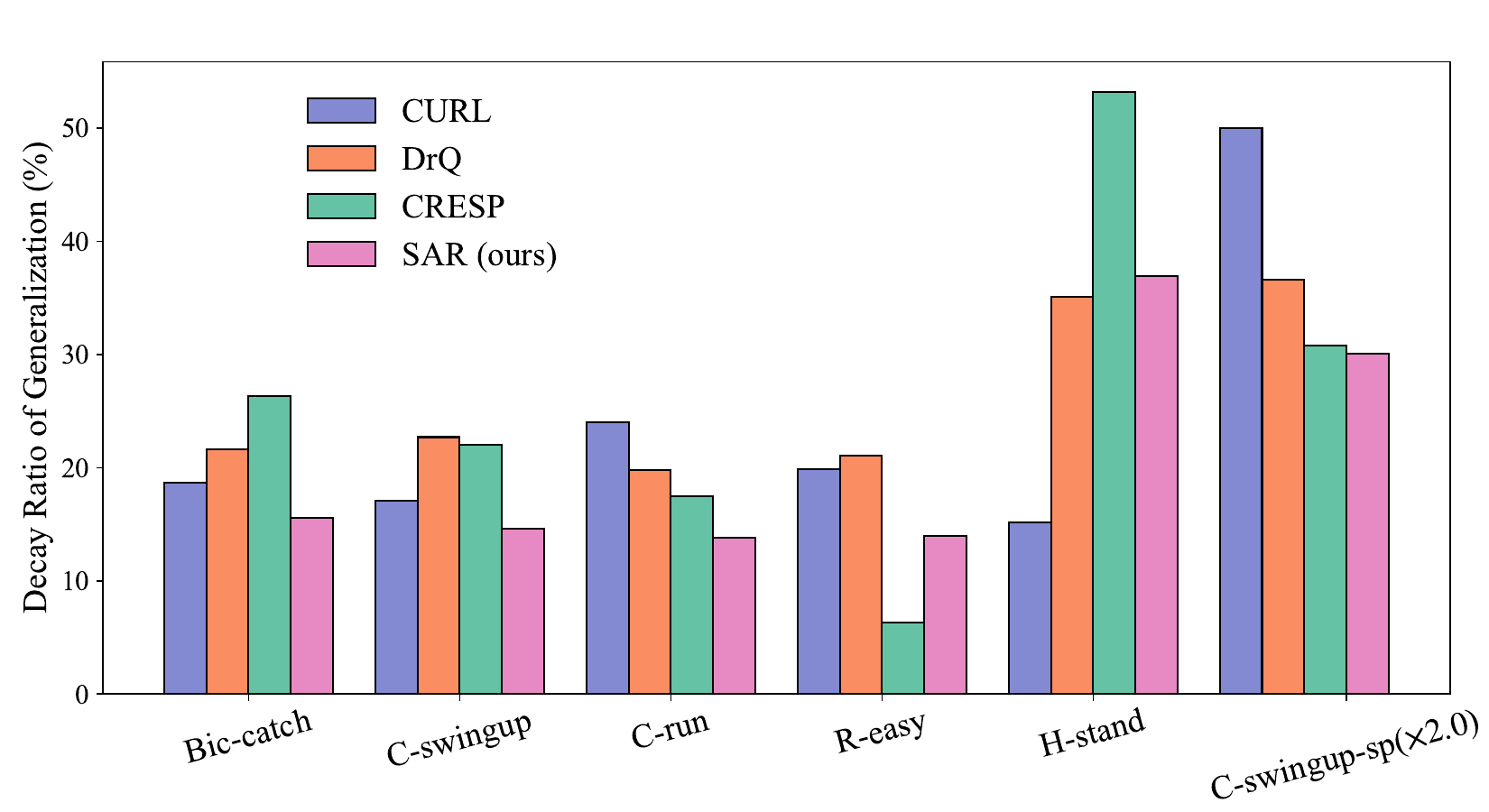}
\caption{Comparison results of generalization decay ratio. It is calculated based on the score results of the models in the training stage with seen distractions and the evaluation stage with unseen distractions. For visualization, we halve the decay ratio value of Cartpole-swingup-sprse.}
\label{fig:5}
\end{figure}
We quantify and analyze the ability of different representations to induce task-relevant information during the training and evaluation stages. Fig. \ref{fig:4} shows the performance comparison curves of baselines during the training stage. In this stage, with iterative training, the distracting backgrounds (videos) will be gradually learned and familiarized by agents, i.e., the distracting backgrounds are seen for agents, so the agents are easier to achieve generalization. In contrast, since the distracting backgrounds used during training and evaluation are independent of each other, i.e., the background videos used during the evaluation stage are not seen during the training stage, some models with weak generalization ability may have significant performance degradation due to the unseen distracting backgrounds. Of course, if the representation learner can sufficiently understand and determine the task-relevant information, the agent should perform equally in the training and evaluation stage.

To quantify the generalization ability of the models, we denote the generalization decay ratio of the model performance from the training (seen background distractions) to evaluation (unseen background distractions)  stage as $\rho=\frac{{score}_{train}-{score}_{evel}}{{score}_{train}}$. The comparison bar charts of the generalization decay ratio for each method on different tasks are shown in Fig. \ref{fig:5}. To be specific, we can summarize the main two conclusions: i) Our SAR method that learns state representation via action sequences achieves the lowest generalization decay ratio compared with the baselines on most tasks, and shows better robustness to the unseen video backgrounds. ii) There is a performance loss when the methods generalize from training to the evaluation stage, and the general decay range of performance is $10\%\sim30\%$, where the largest performance loss is on the complex tasks of Hopper-stand and Cartpole-swingup-sprse.

\subsection{t-SNE Visualization for Latent Spaces}

\begin{figure*}[t]
\centering
\includegraphics[height=4.5cm]{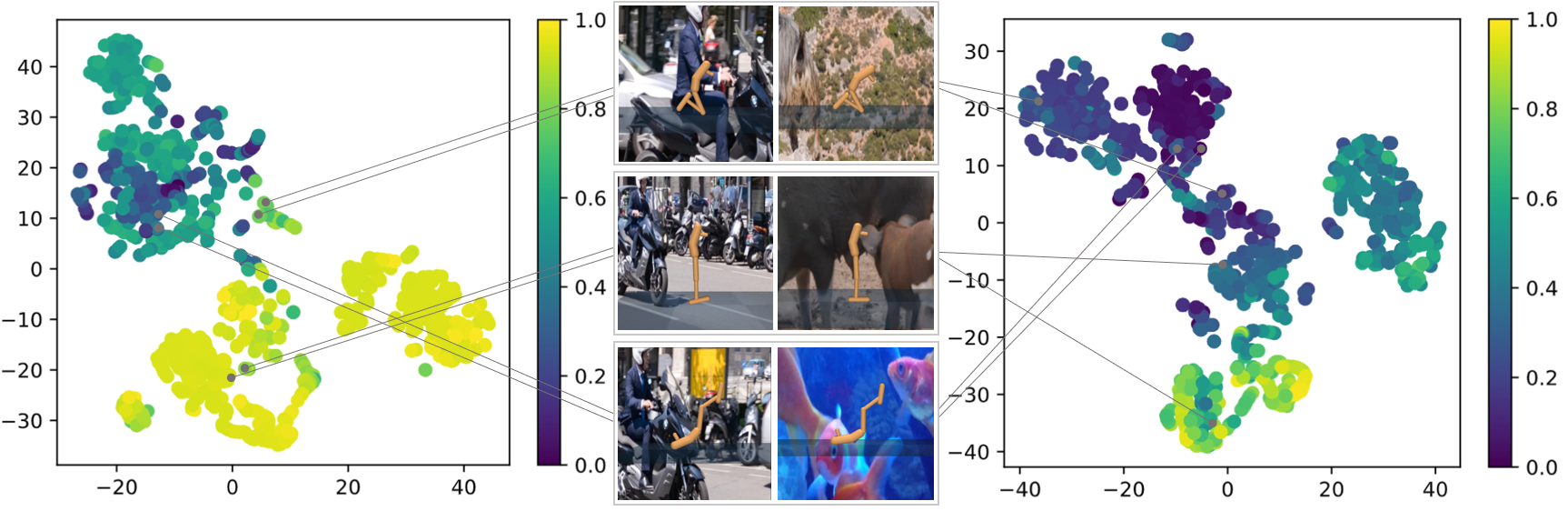}
\caption{t-SNE visualization of latent state spaces learned with converged SAR (left) and DrQ (right). We color-code the embedded points with reward values (higher value yellow, lower value purple). The middle part of the figure is three groups of observations with similar tasks but different background distractions, and the link points represent the projection positions of observation codes under the latent spaces of SAR and DrQ.}
\label{fig:6}
\end{figure*}

We employ the t-SNE visualization to observe the distribution maps of observation transitions under the latent state spaces of SAR and DrQ encoders, where the transitions come from the same batch of 4-episode interaction data, and to establish the relationship between the distribution and the observations with background distractions. In addition, we associate observation images and embedding points according to their sequence numbers annotated in advance. As shown in Fig. \ref{fig:6}, the embedding maps on the left and right correspond to the 2D projections of the observations in the SAR and DrQ latent state spaces, respectively, and the middle shows three groups of random observation samples with similar task information but different video backgrounds. For the projection pairs of these samples, we find that they are very close in the latent state space of our SAR and have similar state values (similar colors). Contrary to SAR, the distances and state values in the latent spaces of the DrQ encoder are far apart, which obviously deviates from reality. In other words, for unseen background distractions, the proposed SAR method can accurately extract task regions through action sequences and compresses them into a low-dimensional representation, which has task and value information that are nearly consistent with the real state. In summary, the t-SNE visualization experiment strongly confirms our previous hypothesis that the method of sequential action-induced invariant representations can better learn representations than baselines.

\subsection{Application in Autonomous Driving}

The inputs of real-world visual control systems such as autonomous driving often contain task-irrelevant elements, such as clouds, mountains, and light, it is necessary to extract as much task-relevant information as possible, e.g., the information about road parts, other vehicles, or obstacles, in order to better autonomous driving under complex and unstructured distractions. To verify the state representation and generalization ability of the algorithms under real visual observations, we applied them to the field of visual autonomous driving based on the CARLA simulator~\cite{dosovitskiy2017carla}.

\begin{figure}[th]
\centering
\includegraphics[height=4.3cm]{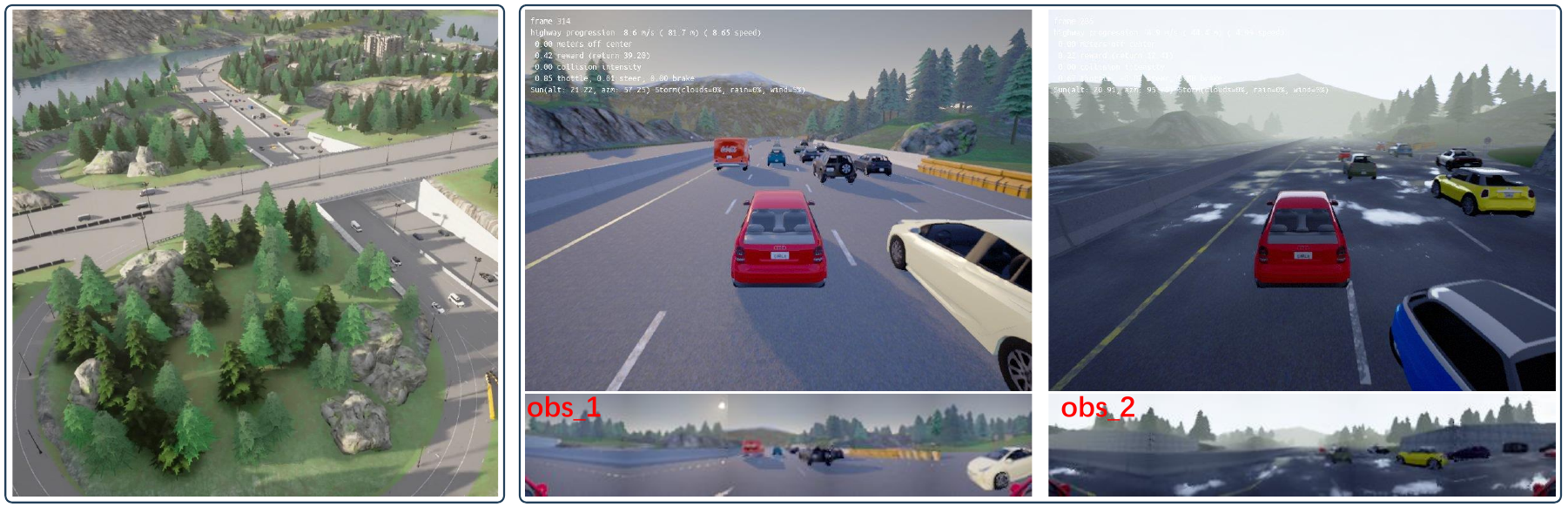}
\caption{\textbf{Left:} the bird's-eye view of the Town04 map. \textbf{Right:} the upper right is two live photos during the training period, in which the red vehicle is the trained agent. Corresponding to the upper scenes, the lower right marked with ``obs\_'' is two wide-angle down-sampled images spliced by 5 cameras arranged on the roof, which are directly utilized as the observation input for the agent training.}
\label{fig:7}
\end{figure}

\textbf{CARLA Task}~CARLA is a powerful autonomous driving simulator that has visual inputs and physical effects nearly the same as the cars in the real world, where we can set rich task components, weather, lighting and NPC vehicles in the scene to meet requirements of different tasks. In this experiment, we choose the officially provided Town04 map with a ring highway containing a crossroad, as shown in Fig. \ref{fig:7}. Concretely, we set up a 300-degree wide-angle image collector composed of 5 cameras with 60-degree views on the roof of the vehicle agent. The agent needs to operate the steering, brake and accelerator to control the vehicle's movement under conditions of wind, rain, cloud or sunlight changes.

\textbf{Task Setting}~ The goal of agents is to safely drive as far as possible in a limited 1000 environmental steps, and the episode will be reset when the agent exceeds the maximum environmental step, goes out of road bounds, or crashes. The observation input for training is an RGB image concatenated by 5 cameras, with a size of $84\times420$ pixels, and the control actions are standardized speed and steer. Following Zhang et al.~\cite{zhang2020learning}, the reward function is set as $r_t(s,a) =\textbf{v}^T \hat{\textbf{u}}_{highway} * \Delta t - C_i*impulse-C_s*|steer|$, where $\hat{\textbf{u}}_{highway}$ is the unit vector of the highway,  $\textbf{v}^T \hat{\textbf{u}}_{highway}$ represents the effective speed which is projected by the vehicle speed onto the highway, and then is multiplied by the unit time $\Delta t$ to represent the effective distance, $impulse$ is the clash force ($N/s$) obtained via physical calculation, and $steel$ is the output amplitude of the steering. Detail parameters can be seen in Table~\ref{tab:3}. Finally, it should be emphasized that learning a basic driving policy through very short training steps (1e5 steps) in such real complex scenarios is a challenging task. 

\begin{figure*}[tb]
\centering
\includegraphics[height=6.0cm]{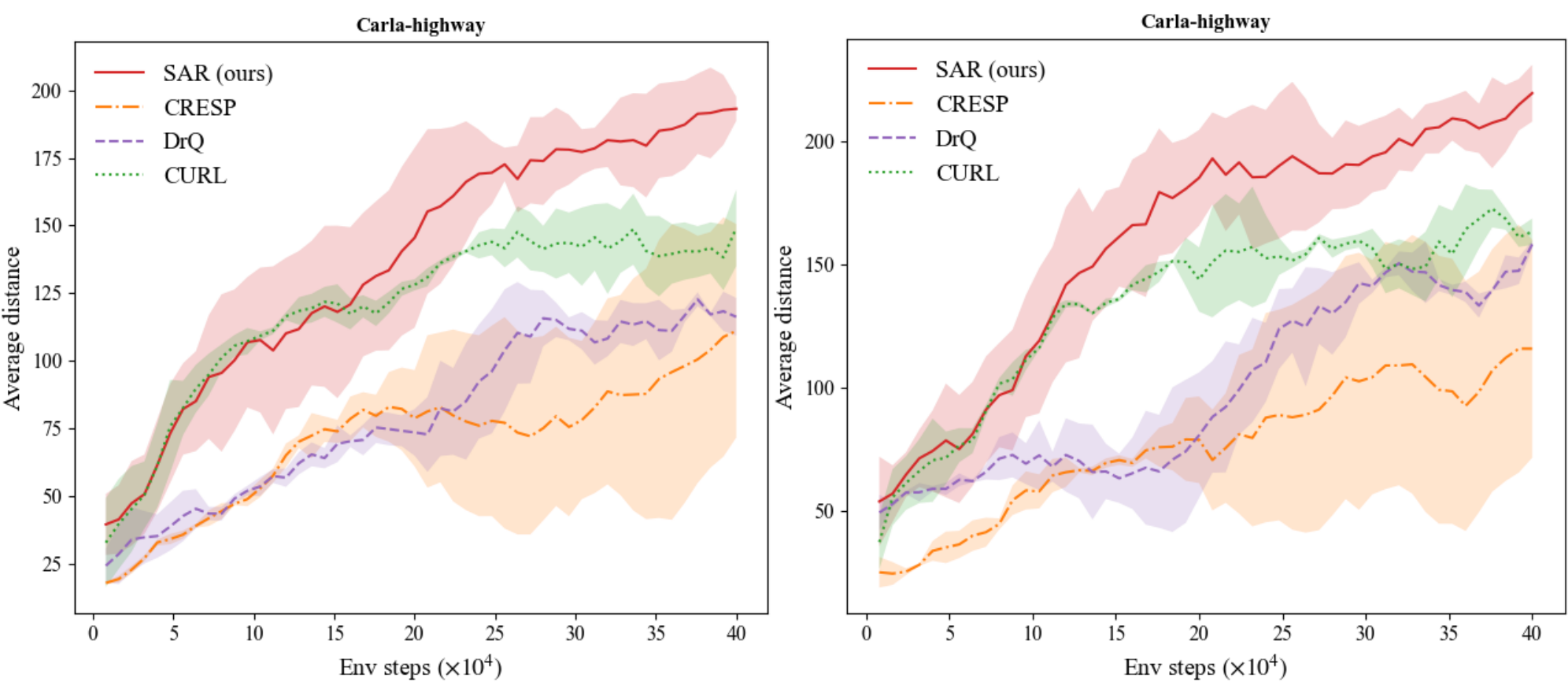}
\caption{Performance comparison of Carla task under 400K environment steps. \textbf{Left:} evaluation curves of average episode reward for SAR and the baselines.\textbf{Right:} average episode driving distance for SAR and the baselines.  Each curve of each method is averaged over 3 seeds and is smoothed for visual clarity, where the shading is the standard deviation. It is worth noting that in this experiment, the DBC algorithm was abandoned by reason of its public code could not reproduce the standard performance.}
\label{fig:8}
\end{figure*}

\begin{table*}[tb]\scriptsize
    \centering
    \caption{Comparison results of the best episode scores on Carla tasks. (mean $\&$ standard deviation for 3 seeds)}
    \label{tab:4}
    \begin{tabular}{lcccccc}
        \toprule
        Methods      & Reward &  Distance~(m) & Successes  episodes & Average steer & Average brake  & Crash intensity \\
        \midrule
        CURL         & 149.1$\pm$14.1  &        172.5$\pm$7.6      &       14\%          &      9.76\%             &   1.75\%             & 4311 \\
        DrQ          & 122.8$\pm$2.58    &   158.1$\pm$1.54   & 21\% & 12.02\%    & 1.66\%    & \textbf{3313} \\
        CRESP        &    111.1$\pm$39.4    &   115.7$\pm$50.2    &     8\%      &   15.53\%   &      1.93      &  3898       \\
        SAR (ours)   & \textbf{193.2$\pm$4.5} & \textbf{219.4$\pm$11.5} &\textbf{ 30\%} & \textbf{7.86\%} & \textbf{1.43\%} & 4023 \\
        \bottomrule
    \end{tabular}
\end{table*}

\textbf{Evaluation}~ We choose CRESP, CURL, and DrQ algorithms as the baselines. As shown in Fig. \ref{fig:8}, our algorithm is significantly better than the baselines in both crucial comparisons of episode reward and episode distance. To be specific, the best average reward and driving distance of the proposed algorithm are 193 $m$ and 219 $m$, which are 29.5\% and 27.3\% higher than the suboptimal baseline respectively, as shown in Table \ref{tab:4}. In addition, although the learning parameters of our algorithm are slightly increased than those of the CURL and DrQ, the convergence rate is still higher than that of all baselines. The successes episode ratio is another important metric, which represents the proportion of episodes that reached the episode horizon. As can be seen from Table \ref{tab:4}, our method also improves by 42.9\% over the suboptimal baseline. Furthermore, for the metrics of the average steering amplitude change, average brake degree, and crash intensity, which represent the stability metrics of the vehicle during driving,  our method outperforms the baselines in terms of the first two metrics, for crash intensity, the performance of our algorithm is lower than the baselines, the possible explanation is as the agents of the baseline travel a shorter distance than ours, so some of the crashes can be avoided. As a whole, the significant advantages of the comparative experiments empirically indicate that the proposed algorithm is able to better handle task-irrelevant distractions, thereby learning an excellent policy.

\section{Related work}

To promote the computation and generalization of downstream decision-making tasks with high-dimensional inputs, it is usually required to learn low-dimensional representations from the high-dimensional images\cite{10078231,9529088,mazoure2021cross,chen2020simple,dong2022identifying,huang2022unified}. Traditional methods including DQN \cite{mnih2015human}, Reformer \cite{kitaev2019reformer}, GMAQN \cite{liang2021gated}, GTrXL \cite{parisotto2020stabilizing} and RAD \cite{laskin2020reinforcement} typically trains a state encoder in an end-to-end manner, but they are difficult to handle non-structural high-dimensional scenarios in reality. In the literature, other usually adopted approaches are as follows. 

\textbf{Reconstruction-based Representations}~By decoupling representation learning from policy learning, pixel-level image reconstruction and latent state encoding are also typically used methods in the early stage of the development of representation learning \cite{fu2021learning,higgins2017darla}. For example, Lange et al. \cite{lange2012autonomous} optimize the reconstruction loss to predict the information required by the policy in a two-stage learning process. Yu et al. \cite{Yu2022MaskbasedLR} and Liu et al. \cite{liumasked} introduce mask-based reconstruction loss, which aims to reconstruct data and facilitate the learning of state representations. Furthermore, model-based methods, such as the PlaNet \cite{hafner2019learning} and SLAC \cite{lee2020stochastic} algorithms, use the reconstruction loss to learn latent state models. However, these methods often require extensive manual tuning, and moreover,  the learned representations can be hardly task-relevant.

	\textbf{Contrastive-based Representations}~Recently,  self-supervised learning of representations in reinforcement learning has attracted much attention, and contrastive representation learning is one of the typical approaches \cite{laskin2020curl}, where the InfoNCE loss \cite{oord2018representation} is optimized to maximize the mutual information between anchors and positive samples while staying away from the information of negative samples \cite{henaff2020data}. To embed task-relevant information into the representation, Liu et al. \cite{liu2020return} and Chen et al. \cite{chen2022hard} divided positive and negative samples by task-related rewards, Laskin et al.\cite{laskin2022cic} used contrastive learning between state transitions and abstract skills to learn behavior representations, Fan et al. \cite{fan2022dribo} introduced temporal information to establish multi-view contrastive learning and Agarwal et al. \cite{agarwal2020contrastive} adopted policy similarity metric (PSM) as the temperature parameter for contrastive learning.

\textbf{Metric-based Representations}~Most of the work on self-supervised learning representations focuses on how to determine a representation that is equivalent to the task. Although some work with contrastive learning has achieved some success, as the related methods mainly optimize a relaxed lower bound of mutual information, the problem of out-of-distribution generalization cannot be handled well. Recently, by directly optimizing the distance between encoding state and MDP elements, the task-related representations via the metrics of task equivalence can be extracted, which have shown better generalization in environments of similar tasks but unfamiliar backgrounds \cite{chenlearning}. The methods mainly include: deep bisimulation metrics (DBC) \cite{zhang2020learning}, which optimizes the encoding distance between samples to make it equal to the sum of corresponding rewards and transitions distance; policy similarity metric (PSM) \cite{agarwal2020contrastive,chen2022learning}, which introduces one-step action distance to measure the similarity of states; return-based contrastive learning \cite{liu2020return}, which uses the return to distinguish the task information and maximize the mutual information between encoder and high-return samples; CRESP \cite{Yang2022LearningTR}, which takes the insight that the same reward should correspond to the same task, then the distance between the predicted distribution and the real distribution of reward sequences is minimized.


Inspired by the aforementioned work, especially the PSM and CRESP methods, in this paper, we also aim to optimize an auxiliary loss to promote task-relevant representation learning. But rather than using the rewards or one-step action, in our method, sequential actions are modeled to capture favorable state representations, which have been shown in experiments that can determine task-relevant representations more accurately.

\section{Conclusion}

We propose the SAR method that leverages action sequences to induce invariant state representations. The SAR is derived from the idea that there is a consistent relationship between real task information in sequential observations and control signals of the sequential actions. By modeling this relationship, our SAR has achieved the ability that quickly lock on the task-relevant information controlled with the sequential actions, which greatly improves the representation performance in unstructured scenes with background distractions.

We compare SAR against strong baselines in the DMControl suite tasks and evaluate it under the challenging setting with unseen background distractions. The results show that SAR achieves the best performance on most of the tasks in the DMC500k benchmark as well as complex tasks with sparse rewards. We also demonstrate again the effectiveness of SAR at disregarding task-irrelevant information by applying it to real-world autonomous driving with natural distractions. Further, we quantitatively analyzed the performance decay of models from the training to the evaluation stage and observed the encoding states of observations under latent spaces, which is still able to well support our conclusions. Finally, the core auxiliary module in the SAR method can be extended straightforwardly to arbitrary visual DRL algorithms.
\begin{table}[H]\normalsize
    \centering
    \caption{Hyperparameters used for experiments of DMControl tasks}\label{tab:2}
    \begin{tabular}{ll}
        \toprule
        Hyperparameter        & Value  \\
        \midrule
        Observation rendering           & 100 $\times$ 100           \\
        Observation downsampling           & 84 $\times$ 84           \\
        Augmentation               & Random shift\\
        Training frames      & 500000           \\
        Replay buffer capacity   & 100000           \\
        Initial exploration steps  & 1000 \\
        Action repeat       & 8  Cartpole-wingup         \\
                            & 4  otherwise        \\
       Stacked frames     & 3           \\
		Evaluation episodes & 10 \\
         Batch size         & 256           \\
          learning rate  & 0.0005           \\
          Discount factor & 0.99 \\	 
         Actor update frequency         & 2           \\
         Critic update frequency         & 2           \\
        Actor log stddev bounds       & [-5,2]           \\
         Init temperature         & 0.1           \\
          Actions sequence length  & 5           \\
          State representation dimension & 50 \\
         Optimizer         & Adam           \\
        \bottomrule
    \end{tabular}
\end{table}

\begin{table}[H]\normalsize
    \centering
    \caption{Partial hyperparameters used for experiments of CARLA tasks}\label{tab:3}
    \begin{tabular}{ll}
        \toprule
        Hyperparameter        & Value  \\
        \midrule
        Camera number                            & 5  \\
        Full fov angles                               &  5$ \times $60 degree   \\
        Observation downsampling           & 84 $\times$ 420           \\
        Initial exploration steps  & 100 \\
        Training frames      & 400000           \\
        Action repeat       & 4          \\
         Batch size         & 128           \\
          $\Delta t$             &   0.05 seconds          \\
         $C_i$              &  0.0001       \\
         $C_s$             &   1.0           \\
        \bottomrule
    \end{tabular}
\end{table}

\section*{Acknowledgments}
We would like to thank the project funds supported by the National Natural Science Foundation of China (No.61772438 and No.61375077).

%


\bibliographystyle{iclr2016_conference}
\bibliography{ref}

\end{document}